# Clinical BioBERT Hyperparameter Optimization using Genetic Algorithm

Navya Martin Kollapally
*Department of Computer Science*
*New Jersey Institue of Technology*
Newark, NJ, USA
nk495@njit.edu

James Geller
*Institute of Data Science*
*New Jersey Institue of Technology*
Newark, NJ, USA
james.geller@njit.edu

*Abstract*— Clinical factors account only for a small portion, about 10-30%, of the controllable factors that affect an individual's health outcomes. The remaining factors include where a person was born and raised, where he/she pursued their education, what their work and family environment is like, etc. These factors are collectively referred to as Social Determinants of Health (SDoH). The majority of SDoH data is recorded in unstructured clinical notes by physicians and practitioners. Recording SDoH data in a structured manner (in an EHR) could greatly benefit from a dedicated ontology of SDoH terms. Our research focuses on extracting sentences from clinical notes, making use of such an SDoH ontology (called SOHO) to provide appropriate concepts. We utilize recent advancements in Deep Learning to optimize the hyperparameters of a Clinical BioBERT model for SDoH text. A genetic algorithm-based hyperparameter tuning regimen was implemented to identify optimal parameter settings. To implement a complete classifier, we pipelined Clinical BioBERT with two subsequent linear layers and two dropout layers. The output predicts whether a text fragment describes an SDoH issue of the patient. We compared the AdamW, Adafactor, and LAMB optimizers. In our experiments AdamW outperformed the others in terms of accuracy.

## I. Introduction

Social determinants of health (SDoH) are the non-clinical factors such as where an individual is born, lives, studies, works, plays, etc. that affect a wide range of clinical outcome [1]. SDoH contribute to socially influenced diseases including obesity, diabetes, depression, respiratory illnesses, cardiovascular diseases, etc. There is a complex "interaction and feedback loop" operating among SDoH [2]. For example, if a person is forced to continuously work more than 70 hours per week just to make ends meet, s/he may not have time for doctors' visits, let alone for social interactions and cooking nutritious food. S/he may be dwelling in poor housing conditions close to a source of air pollution. All these factors affect a patient's medical outcomes. In case of an emergency or hospital admission caused by work related stress or other factors, s/he might be pushed towards or across the edge of financial viability. Existing research has indicated that most of the SDoH data in Electronic Health Records (EHRs) are represented as unstructured text [1, 3]. We hypothesize that SDoH data is mostly recorded as text, because of the lack of clinical codes to record SDoH concepts [4] as structured data. Another reason for the dearth of SDoH data being recorded is that "*Asking busy staff to collect additional information is burdensome*", according to Dixon [4].

A common method of collecting SDoH is to request patients to fill out a questionnaire during physician visits. A more advanced method is to link EHR data with SDoH data extracted by analyzing state and county level census data [5]. This approach has been taken by the Methodist Healthcare Ministries of South Texas and the statewide Health Information Exchange (HIE) for San Antonio. In the absence of a widely accepted method of collecting SDoH data, optimized techniques for extracting sentences or phrases relevant to SDoH are necessary.

Ontologies play an important role in clinical text mining. Medical Ontologies/terminologies are used to identify and extract information from clinical documents. The UMLS Metathesaurus [6] is a large biomedical resource that includes standard biomedical vocabularies like SNOMED CT [7], ICD-10-CM [8], MeSH [9], MedDRA [10], etc. Researchers have developed many algorithm-based techniques that extract semantic and entity-based information using the UMLS [6]. Tools such as QuickUMLS [11] and MetaMap [12] have been developed for medical concept extraction. These tools work well for concept-level extraction in the baseline model, but are unable to provide a good recall value for phrase-level extraction that carries the context information [3]. Many user-generated phrases like 'verbally responsive', 'vitals stable on admission' and 'unresponsive patient with abnormal vitals' that clinicians use daily may not be captured at the granularity required using only concepts from the UMLS. Hence, we are utilizing concepts from our specialized SOHO ontology [13] along with regular expression (regex)-based techniques for identifying relevant text.

We are utilizing a deep neural network BERT-based model for clinical note classification. The performance of a machine learning model depends on the quality of data it is trained with, but an equally important factor is the correct choice of hyperparameters. There are various methods to identify optimal hyperparameters of a model. They include *Bayesian optimization* [14], *grid search* [15], *evolutionary optimization* [16], *meta learning* [17] and *bandit-based methods* [18]. All these techniques have a search space defined by the choice and range of parameters under consideration. In this work, we are deriving optimized hyperparameters for a neural network model using a genetic algorithm. We considered three base optimizers, namely AdamW [19], Adafactor [20] and LAMB [21]. We use a genetic algorithm with n-bit crossover and random bit flip mutations to obtain the optimal candidate solution.

The goal of this paper is to identify, from a large set of clinical text samples (MIMIC III database [22]), text that express an SDoH sentiment about the described patient. This is achieved in a two-step process. First, we extract text samples with a *regular expression*, which is looking for concepts from the SOHO SDoH ontology in the input text. However, some text samples use a SOHO ontology term in an incidental way, not really referring to an SDoH issue of the patient. In order to classify text input as SDoH text or not, we use a neural network pipeline. We combine a Clinical BioBERT [23] model with a *neural network classifier layer*. We also optimize the selected hyperparameters of Clinical BioBERT using a genetic algorithm to achieve higher accuracy of classification.

## II. RELATED WORK

Patra et al. [24] have published an extensive review paper on SDoH data extraction using NLP techniques. The authors identified 6,402 candidate publications from the ACL Anthology, PubMed, and Scopus. Only 82 papers were included by the authors in the final analysis. The authors state that the initial step in SDoH extraction from EHRs is creating a lexicon of SDoH categories and the second step is developing a rule-based supervised system to locate these concepts in EHR text. Only seven of the cited studies were using neural networks as the major mechanism for the NLP task.

Han et al. [25] developed a framework for multilabel classification of text data from the "social work" category in MIMIC III clinical notes. They used only terms from SNOMED CT and DSM-IV. The authors utilized three deep neural networks for processing: LSTM, BERT and CNN. They concluded that the BERT model outperformed LSTM and CNN. BERT was most effective in classifying text from the "occupational" category and least effectively identified text from the "Non SDoH category," according to the authors. For multilabel classification the authors obtained an F1 score of 0.774 for the "occupational" category. The authors used the Adam optimizer with momentum set to 0.9, learning rate equal to 1e-04, and number of epochs=10. Since the performance metric of multiclass classification cannot be compared with the metric for binary classification, we have trained a BERT model with the architecture and hyperparameters mentioned in Han's paper to compare the performance with our optimal parameters. In Section VIII, we will present this comparison.

Ahsan et al. [26] have created a publicly available dataset with 7,025 discharge summary notes for seven social and behavioral determinants of health. The authors extracted only the text from the social history section of discharge notes using MedSpacy [27] and it was human annotated. They utilized the Clinical BioBERT [23] model for classification. They mentioned the lack of hyperparameter tuning of the transformer model as a limitation.

Lituiev et al. [28] used NLP methods to extract adverse SDoH information for patients with chronic lower back pain. The authors performed an offline knowledge extraction, i.e., without the need for a clinician for immediate decision-making with a pre-defined ontology. The authors utilized progress notes, history and physical (H&P) notes, emergency department (ED) provider notes, patient instructions, and telephone encounters from the University of California, San Francisco (UCSF) data warehouse to extract SDoH data corresponding to the seven SDoH categories housing, food transportation, finances, insurance coverage, marital and partnership status and social support. Their ontology is based on SNOMED CT with an additional 68 second level and 10 first level classes. The notes were annotated by four annotators and the authors utilized a RoBERTA evaluation-based model for the named entity recognition (NER) task.

## III. BACKGROUND

Ontologies help with defining concepts, relationships between them, and instances that can be utilized in biomedical research and applications [29]. BioPortal, maintained by the National Centre for Biomedical Ontology (NCBO) [30] is a repository of over 1000 biomedical ontologies. SOHO is a Social Determinants of Health ontology available in BioPortal [31] that we developed in previous research. SOHO has a root class "Social determinants" and two children "of_health" and "of_health equity" (using a shortened term by omitting "Social determinants"). The SOHO ontology consists of nine child classes under Social determinants → of health. They are "Economic stability," "Education access and quality," "Healthcare access and quality," "Neighborhood and built-in environment," "Social and community context," etc. Fig. 1 shows a Protégé visualization of a part of SOHO. The SOHO ontology consists of 189 classes and 585 axioms of which 207 are logical axioms.

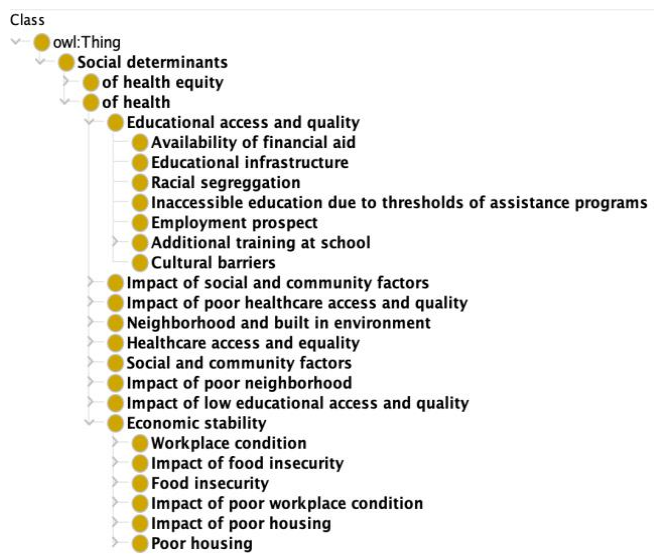

Fig. 1. Snippet of SOHO from Protégé

Natural language processing (NLP) techniques are used to extract information from text documents. NLP techniques are widely used to extract SDoH data from EHRs. BERT (Bidirectional Encoder Representation from Transformers) transformer-based models revolutionized the NLP field [32]. Each encoder layer in BERT has a self-attention head and a feed forward neural network. BERT was pretrained on 2.5 B words of Wikipedia and 800 M words from Google's Book corpus. For every input word embedding, the self-attention head computes query vectors, a key vector and a value vector. As the word vectors pass through, each encoder layer adds its own attention score to each word representation. BioBERT is based on the BERT base model with 12 encoded layers, 789 hidden states and 110M parameters. BioBERT [23] was trained on 4.5B PubMed documents and 13.5B PMC articles. BioBERT is fine-tuned by using 2 million notes from MIMIC-III to generate the Clinical BioBERT model [23]. Fig. 2 represents the evolution of Clinical BioBERT from BERT.

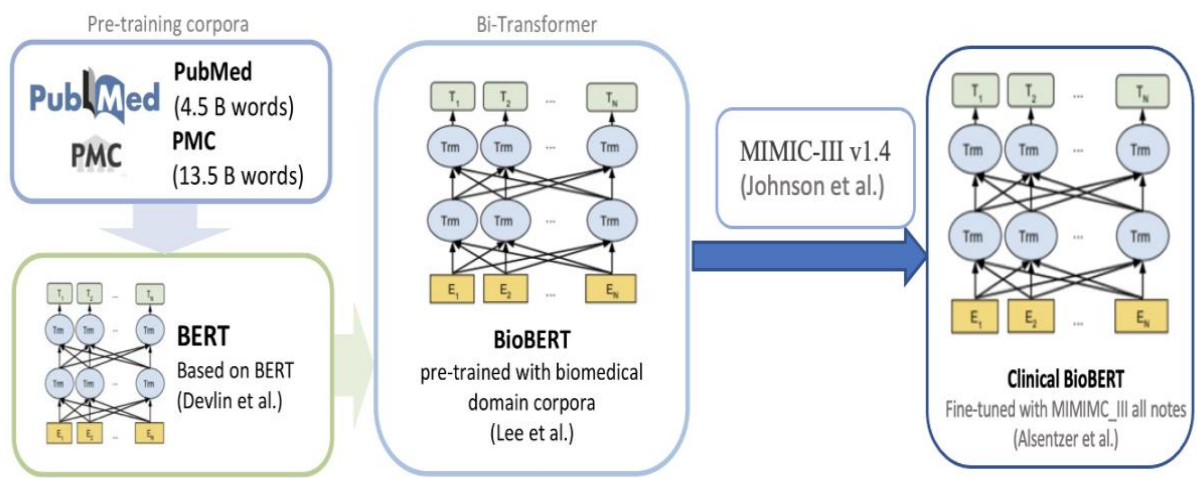

Fig. 2. From BERT to Clinical BioBERT (modified from [23])

The BERT framework involves pre-training and fine tuning. In the pre-training phase the input consist of word embeddings augmented with special tokens [CLS] and [SEP]. [CLS] is a special classification token. The last hidden state of BERT corresponding to the [CLS] token will be used for the classification step. [SEP] is a separator token. In pre-training, the input is a sequence consisting of the special token [CLS], tokens of a sentence A, a special token [SEP], and tokens of another sentence B that follows A in the corpus [23]. Each token's initial embedding encodes its content, its position in the sequence, and the sentence it belongs to in the corpus. Pre-training is conducted by minimizing losses for two tasks, Masked Language Modelling, which predicts tokens that are randomly masked, and Next Sentence Prediction, which predicts whether sentence B follows A or not. Given a set of paragraphs with positive and negative labels representing whether the paragraph is in SDoH context or not, we fine-tune a pre-trained Clinical BioBERT model along with a downstream binary classifier on the cross-entropy loss. The inputs of Clinical BioBERT are the tokenized vectors with the maximum length set to 128. The classifier consists of two linear layers with dropout layers that takes as input the embedding of [CLS] tokens from Clinical BioBERT's last-layer and transforms it into a 2-dimensional vector before applying to the output linear layer for sentence classification.

The role of an optimizer in a neural network model is to find the accurate combination of weights such that the error when mapping inputs to outputs becomes minimal [33]. Adaptive optimization algorithms such as Adam have a better performance compared to Stochastic Gradient Descent (SGD) optimization [19] in some scenarios. The learning rate of SGD as a hyperparameter is difficult to finetune, since the magnitudes of different parameters vary widely in the training process. AdaGrad, Adam, and RMSProp are various adaptive variants of SGD [19]. Adaptive optimizers like Adam initially gained a lot of popularity, but later it was found that adaptive optimizers with L2 regularization did not work as well as Stochastic Gradient with Momentum. An improved version of Adam called AdamW [34] was developed, which exhibits a better performance compared to Stochastic Gradient with Momentum for most datasets. This improvement is achieved by decoupling the weight decay from optimization steps taken with respect to the loss function [34].

The adaptive gradient optimizers have a high memory requirement. AdamW must keep track of first momentums and second momentums to calculate weight decay or gradient, thus tripling the memory requirements. Adafactor [20] guarantees the same empirical performance as AdamW, but with reduced memory usage. Adafactor drops the first momentum completely and instead tracks the moving averages of the row and column sums of the squared gradients for matrix-valued variables. Hence for an n × m matrix, this reduces the memory requirements from $O(nm)$ to $O(n + m)$. Adafactor can converge without momentum by increasing the decay rate with time and clipping the gradient update [20].

The Layer wise Adaptive Moments optimizer for Batch training (LAMB) uses an accurate layer wise trust ratio to adjust the Adam optimizer's learning rate. As the batch size grows, the number of iterations per epoch decreases [21]. To converge at the same time, we increase the learning rate. Learning rate scaling heuristics (adjusting the learning rate according to a pre-defined schedule) with the batch size do not hold across all problems or across all batch sizes [21]. Learning rate warmups were introduced, but they only helped up to a certain point, at which the learning would start to diverge. Hence LAMB optimizers were introduced with a layer-wise adaptation strategy to accelerate training of neural networks. During local or global minima, the gradient is relatively small, and weights are high. This adds a scaling factor to the learning rate, which prevents getting trapped in local optima.

As networks deepen, it becomes important to achieve zero mean, unit variance (ZMUV) weights. At each layer, these weights are multiplied together, so if the norm of the weight matrix diverges from ZMUV, the values of the weights may explode or vanish. When values of weights are small, the algorithm takes small steps. When values of weights are large, it takes bigger steps. At the beginning of training, layers are supposed to output ZMUV and steps are likely to be small. In contrast, the change in layer gradients will probably be large, since when the algorithm is initially far away from the optimum, the gradients are large. In this way the algorithm for LAMB performs natural warm up as the weights increase. As we approach negligible loss, the gradients will be small, so the trust ratio will keep the learning rate higher than without the trust ratio, thus avoiding local optima.

## IV. MODEL ARCHITECTURE

The Clinical BioBERT model architecture is a multi-layer bidirectional transformer encoder implementation as in [23, 32]. The clinical notes present in MIMIIC III are fed into a regex-based Python script to extract text fragments with SDoH concepts in them. The SOHO ontology is used in this task as shown in Fig. 3. Not all rows of data returned by Python regex script express a strong SDoH sentiment. Hence, a manual review of a subset of approximately 1500 rows of extracted text was performed and 1054 rows were annotated with the label "1" for training the Clinical BioBERT architecture. Negative training samples (1130 rows) were mostly extracted from admission labs, discharge labs and discharge instructions. The resulting 2184 rows of data were split into 80% training and 20% test data. The 80% were again split 80:20 and the resulting 64% were used for training with 16% for validation. The input data is converted to token embeddings, each as a 768-dimensional vector representation. The sum of positional, segment and token embeddings constitutes the input embedding. The input embeddings are first passed through a multi-head self-attention mechanism. The self-attention mechanism generates a set of attention weights that are used to weigh the importance of each token in the input sequence. The context vector is passed through a position-wise feed-forward neural network, which further transforms the context vector. The classification layer takes the CLS token of the last layer and predicts the context of the text sample. The classification layer is made up of two linear layers separated by two drop out layers.

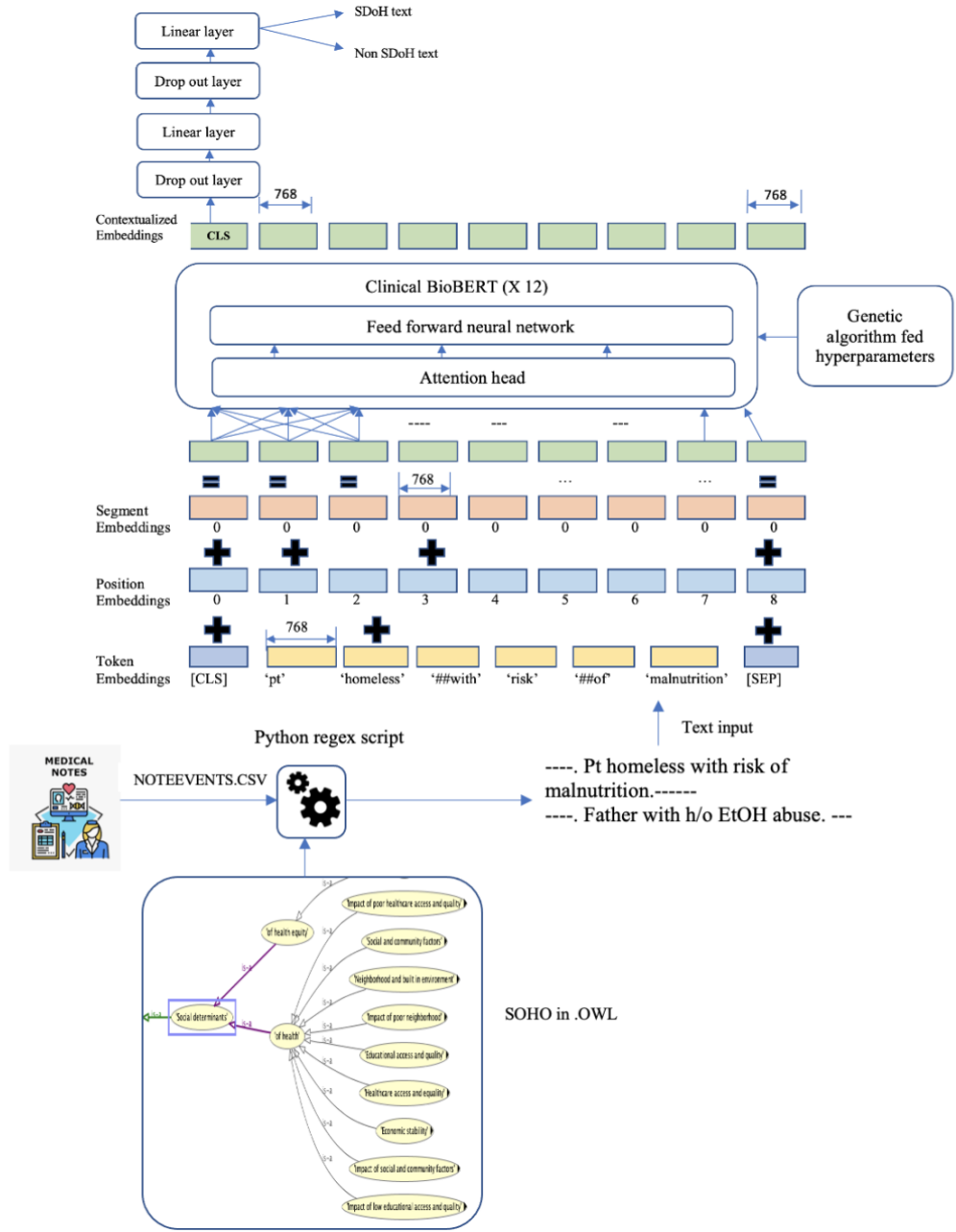

Fig. 3. Model architecture of Clinical BioBERT for SDoH text classification.

## V. EXPERIMENTAL APPROACH

### A. Dataset

We have utilised our SOHO ontology, available in BioPortal, as a reference terminology for extracting concepts from MIMIC III v1.4 [22]. The concepts in the branch "Social determinants of health" were only used for concept extraction from MIMIC III clinical notes. MIMIC-III contains data associated with 53,423 distinct hospital admissions for patients 16 year and up admitted to critical care units between 2001 and 2012 [22]. It also contains data for 7,870 neonates admitted between 2001 and 2008 [22]. The data covers 38,597 distinct adult patients and 49,785 hospital admissions. We specifically utilized clinical notes available in the NOTEEVENTS table which is a 4GB data file. The file contains nursing and physician summaries, ECG reports, radiology reports and discharge summaries. The gold standard for extraction is derived by the SDoH concepts present in our SOHO ontology [13]. The input text is tokenized and

To extract SDoH-related sentences, we used regex regular expressions. We started with concepts in the leaf nodes of the SOHO ontology and moved up its DAG (Directed Acyclic Graph) structure. Using regex expressions, when we found matching concept classes from NOTEEVENT files, for each sentence, we extracted the preceding four sentences and the succeeding four sentences, since BERT models base their classification on the context of data. We observed in a sample that in MIMIC III clinical notes, SDoH data per patient spanned at most four sentences. Due to the limitations of manual annotation, we used 2184 rows of text for this phase of research.

The NLTK library was used for text preprocessing. Extracted sentences were stripped of special characters and URLs. After stop word removal and converting the text to all lower case, we labelled the dataset as 1054 rows with paragraphs of sentences in each row fitting in the SDoH context with the label 1. We labeled the remaining 1130 rows of paragraphs as 0, because they were not relevant in the SDoH context. A sample from the dataset that has a class label 1 is:

*"Case Management spoke with the patient's long-term nurse practitioner, who stated that at baseline he normally uses a wheelchair and occasionally walks on his prosthesis. Additionally, the nurse practitioner stated that the patient has chronic **drug abuse**, both prescriptions and illicit, and advised not to give the patient any pain prescriptions upon discharge as he had more than enough at home. It was also discussed that the patient would often try to prolong his hospital courses in the past to get more narcotics".*

As seen in the sample text above, which was classified as SDoH-related text, the concept used for mapping from SOHO is "drug abuse." The context of SDoH is limited to just 1 to 4 sentences.

### B. Clinical BioBERT

Both the Clinical BioBERT tokenizer and the pre-trained model [35] were used for transfer learning with SDoH data. More precisely, the word embeddings output from the Clinical BioBERT tokenizer were converted to tensor objects to fine tune the model. To account for co-adaptation, the output from BERT is passed through two dropout layers with drop out probabilities 0.25 and 0.3. The dropout layers are separated by a linear layer with 768 hidden states of a feed forward network for binary classification. At the output there is a fully connected layer, since it is needed to calculate cross entropy loss.

### C. Hyperparameter optimization

Fine tuning an existing transformer model is not time consuming, but finding optimal hyperparameters for the model is a resource intensive task [36]. There are many methods of hyperparameter optimization, such as grid search, Bayesian optimization, and population-based approaches. We employ a genetic algorithm to implement population based hyperparameter optimization. Once optimal hyperparameters have been reached, we train the model on the training dataset to obtain the optimal accuracy with minimal cross entropy loss.

The hyperparameters chosen for this phase of research are optimizer type, epoch number, learning rate ($\eta$) and epsilon ($\varepsilon$). The three optimizers of choice are AdamW, Adafactor and LAMB.

Epoch counts chosen were 5, 10, 15, 20, 25, 35, and 50. (An epoch denotes the number of times the model has seen the entire dataset during training.) The learning rates range from 2e-8 to 1e-1. The learning rate is defined as the pace with which a model updates it parameters or learns the parameter values. Epsilon is a parameter that is added in the square root of the second momentum to avoid a division by zero. Epsilon's range set is from 1e-8 to 1e-4. All these parameter ranges where selected based on the established benchmarks by previous scholarly articles [14, 18, 24].

### D. Evolutionary strartegies

Following the terminology of genetic programming, each of the parameters is encoded as a chromosome using binary encoding. Each chromosome consists of four genes and is 24 bits long. We used two bits to represent the optimizer, six bits for the epoch number, eight bits for the learning rate, and eight bits for epsilon (Fig. 4). We started with a random initial population of 20 chromosomes per generation. The fitness of a candidate solution is based on the accuracy of the model. After implementing n-point crossover and random bit flip mutations (Figs. 5 and 6), the accuracy is evaluated. Selection to the next generation is based on a Roulette wheel approach.

Roulette wheel selection [37] is a probabilistic approach that ensures that the population does not just consist of elite candidates; it also contains some weak solutions. Roulette wheel selection ensures diversity in the selection process, thus reducing the chance of getting stuck at a local optimum in a multimodal problem. Three iterations were performed with 25 population updates in each. The number of generations was fixed as 25 based on the convergence of cross entropy between consecutive iterations.

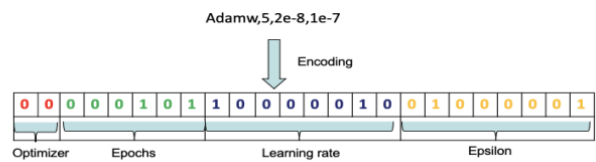

Fig. 4. 24-bit representation of a generation

To perform recombination and mutation operations, we used n-point crossover and random bit flip mutation. Fig. 5 shows a 1-point crossover operation where the crossover happens at the 7-th locus position. At this point the tail from parent B combines with the head of Parent A to generate child 1. To generate child 2, the head of parent B combines with the tail of parent A. We have used a crossover probability ($P_c$) of 0.75. The value 0.75 means that 75% of the generation undergoes the crossover. If the crossover probability is 1 then all the candidates in the generation undergo recombination operation. A value of zero indicates there is no evolution of off-spring. Recombination operations ensure that the best features are likely to persist into the next generation.

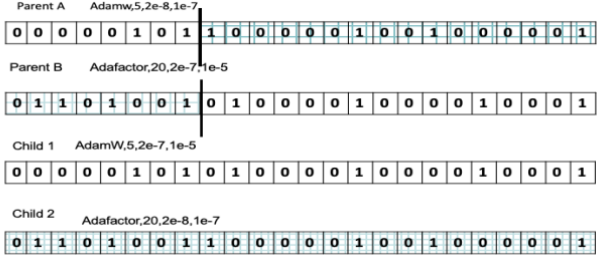

Fig. 5. 1-point crossover to generate input for the next generation

Fig. 6 shows an example of offspring undergoing a random bit flip mutation. Mutations are a way of introducing new features into the existing population. The mutation probability $P_m$ is 0.03 in our analysis. Each bit in a chromosome is considered for a possible mutation by generating a random number between zero and one and if this number is less than or equal to the given mutation probability 0.03 then the bit value is changed. The offspring in Fig.6 is generated by flipping the bits at loci 0, 3, 7, 18, 19. We only choose viable offspring for the next stage, while catastrophic offspring is eliminated.

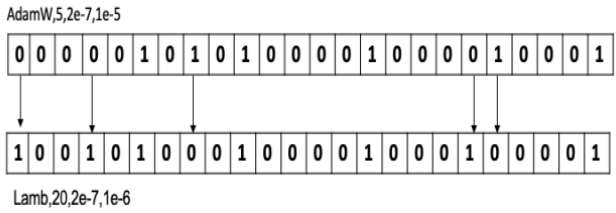

Fig. 6. Bit flip mutation to do an exploitation analysis in the neighborhood

### E. Fitness evaluation

The evolutionary algorithm is guided by a fitness evaluation representing the user's objectives. Thus, the formulation of an ideal fitness function is task specific. For our model, we consider five-fold cross validated accuracy as the fitness function. Accuracy is defined as the ratio of number of correctly classified datapoints to the total number of data instances.

$$Accuracy = \frac{TN + TP}{(TN + FN + TP + FP)} \quad (1)$$

TABLE 1. CONFUSION MATRIX

|  |  | Actual value | |
|---|---|---|---|
|  |  | Positive | Negative |
| Predicted value | Positive | True Positive (TP) | False Positive (FP) |
|  | Negative | False Negative (FN) | True Negative (TN) |

In equation (1) above, TN (True Negative) is the number of negative labels correctly identified as negative by the model and TP is the number of positive labels identified as positive by the model. Similarly, FN (False Negative) are those labels classified as negative, but they have a positive label and FP is the number of data points classified as positive, but they are in fact negative. Table 1 shows the confusion matrix. Results obtained during training and validation are used to guide the search process. We are using the standard approach of reserving a part of the training set for validation, to obtain the optimized data-dependent hyperparameters and a second part for testing. As mentioned, there are two selection mechanisms, the parent chromosome selection and survivor selection to the next generation based on a Roulette wheel.

Fig. 7 represents how we performed the genetic algorithm based hyperparameter optimization. The decoded chromosome values corresponding to valid choices are used as hyperparameters in training of Clinical BioBERT. The fitness of the model is evaluated in terms of accuracy and those hyperparameters corresponding to Roulette wheel selected chromosomes move to next generation. In all three iterations, the stopping criterion of the, accuracy not improving after four successive generations was used. For ease of analysis, we performed 25 generational updates in each iteration even though some iterations converged before 25 updates.

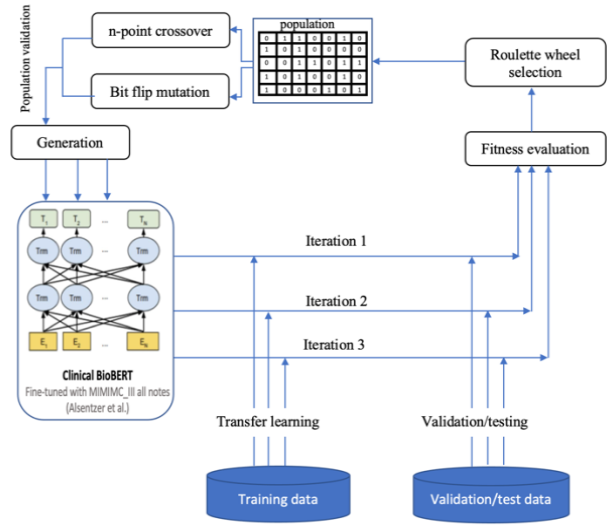

Fig. 7. Evolutionary development of Clinical BioBERT for hyperparameter optimization

## VI. ALGORITHM

Optimization is the process of finding solutions that satisfy the given constrains and meet specific goals at their optimal value. In this algorithm, we are optimizing the set of hyperparameters in Clinical BioBERT such that the cross-entropy loss is minimal and fitness in terms of accuracy is highest. In step 3 selected_chromosome is a list of chromosomes that survives the selection process and counter is a variable maintained to escape local optima. In step 5 elite_acc$^{prev}$ is the accuracy of the best candidate from the previous generation, and elite_error$^{prev}$ is the cross-entropy loss of the best candidate from the previous generation in step 6. In step 7 elitist_acc and elitist_error are the cross entropy loss and accuracy of the best candidate in the current generation. In step 8 max_gen, is the maximum number of generational update in an iteration. Steps 9-21 are the core of the genetic algorithm. It starts with choosing chromosomes with viable combinations of traits, followed by limiting the size of each generation to 20. The first 2 bits encode the optimizer type. We use 00 for AdamW, 01 for Adafactor, and 10 for the LAMB optimizer. Suppose one of the random chromosomes generated has as first two bits 11, which is not a meaningful encoding. Hence those chromosomes will be discarded.

To incorporate the fact that best traits from parents should persist in the offspring we perform n-bit crossover with probability $P_c$. To introduce new traits the chromosomes undergo bit flip mutation with probability $P_m$. We evaluate the fitness of the generation and spin the Roulette wheel 20 times to choose survivors to the next generation. The constraints on the maximum number of generations are 1) 1000 evolutions have passed and the algorithm is not converging toward a global solution; 2) If the accuracy between successive generations stays the same, we continue the process for four generations and if it is not improving it might be stuck in a local optimum or it found the best global solution.

To validate the same, we add more weak chromosomes to ensure diversity, based on the principle of simulated annealing [38] (step 23-31). The variable $acc_g$ is the accuracy and $cel_g$ is the cross-entropy loss of the current chromosome under consideration. We capture the best accuracy in the current run to elitist_acc by comparing the best accuracy so far with the accuracy in each generation. At the end of each iteration the output will have the accuracy and the hyperparameters (encoded in binary) as output for the best candidate in each iteration until convergence is achieved.

| | Algorithm for finding optimal parameter set for Clinical BioBERT |
|---|---|
| 1 | for iteration i=1 to 3: |
| 2 | //start with 24 bit encoded chromosomes, let n be the total number of chromosomes<br>create a set of n random chromosomes $C_1$ to $C_n$ |
| 3 | //list initialization to store the survivor chromosomes<br>selected_chromosome=[] |
| 4 | counter=0 |
| 5 | //variable elite_accprev is the accuracy from best candidate of previous generation<br>elite_acc$^{prev}$=0 |
| 6 | //variable elite_errorprev is the cross entropy loss of best candidate of previous generation<br>elite_error$^{prev}$=0 |
| 7 | //variable elite_accprev is the accuracy from best candidate of previous generation<br>elitist_acc=0 |
| 8 | //counter to record the number of generational updates<br>max_gen=0 |
| 9 | //start of genetic algorithm<br>begin: |
| 10 | //number of generational update is captured<br>max_gen +=1 |
| 11 | //iterating through n-chromosomes<br>for k=1 to n: |
| 12 | validate viable chromosomes |
| 13 | //only valid chromosomes are captured in the list and undergo recombination and mutation<br>selected_chromosome.append($C_k$) |
| 14 | If len (selected_chromosome )=20: |
| 15 | break |
| 16 | apply n point crossover($p_c$) -> selected_chromosome |
| 17 | apply random bit flip mutation($p_m$) -> selected_chromosome |
| 18 | //P include the viable chromosome and offspring<br>let P be the new population with parents and offspring |
| 19 | for g = 1 to len(P): |
| 20 | //decode the chromosome and run Clinical BioBERT model with hyperparameters<br>evaluate the fitness of chromosomes $P_g$ |
| 21 | apply Roulette wheel probabilistic selection and choose 20 new candidates |
| 22 | for j= 1 to 20: |
| 23 | //$acc_g$ is the accuracy from survivor chromosome<br>if $acc_g$ > elitist_acc: |

```
24              elitist_acc = acc_g
25          else if elitist_acc - elitist_acc^prev ~ 0:
26              //to make sure not stuck in local optima we add weak chromosomes
                add diverse valid weak chromosomes to generation
27              counter += 1
28              continue to step 17
29          elitist_acc^prev = elitist_acc
30          elitist_error^prev = elitist_error
31      Continue to step 16 till counter < 5 or max_iter < 1000
32  end:
```

## VII. RESULTS

For the population-based optimization, to find the best global solution, a considerable size population with diversity is a key factor. Our experimental result for each iteration had 25 generational updates, each with a population size of 20. We performed a generational population strategy by which the entire generation is replaced by the newly selected chromosomes. Hence in each iteration we had a total population size of 20*25=500 chromosomes. To avoid the problem of local optima, we considered three different initial configurations, each with 500 chromosomes, thus totaling 500*3=1500 evaluations to derive the best hyperparameters for Clinical BioBERT in the context of SDoH.

During the initial iterations, when the accuracy was not changing in four iterations, we intentionally added some weak solutions with lower accuracy to make sure that the algorithm is not stuck in a local optimum. Roulette wheel selection also guided us to develop enough diversity in the population. The graph in Fig. 8 shows the validation vs training loss curve for iterations with respect to the three optimizers. When we observed significant overfitting, represented by increasing validation loss with decreasing training loss, we adjusted the drop out value to avoid overfitting. The best hyperparameters for the Clinical BioBERT model trained on SDoH data are using the AdamW optimizer with learning rate=2e-8, number of epochs=10, and epsilon=1e-08 from the transformers library in Python, implemented along with a linear warmup scheduler.

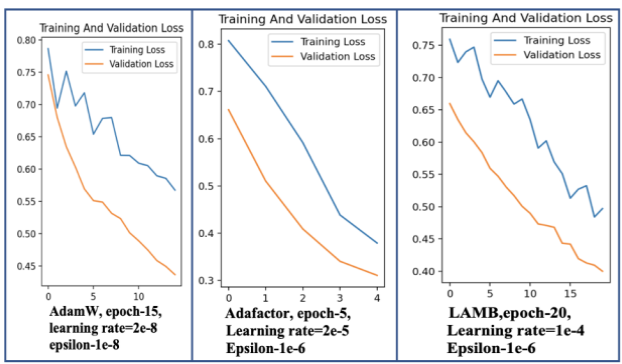

Fig. 8. Sample training vs validation loss curves for three optimizers

Following graph (Fig.9) represents the fitness value of best candidate in each generation plotted for all three iterations. During the iteration whenever a steady accuracy is obtained for successive generation, we used a simple heuristic and added around 25% of weak solutions along with 75% of elitist solutions. This approach often helped us surpass the problem of trapping in local optima. In our iterations whenever the diversity of candidates is maintained the problem of local optima was surpassed and ensured our model converges at global best parameter set. In all three iteration a global best solution was identified by 11 the generation, thereafter there was no improvement in accuracy, even if weak solutions where explicitly added to the generation.

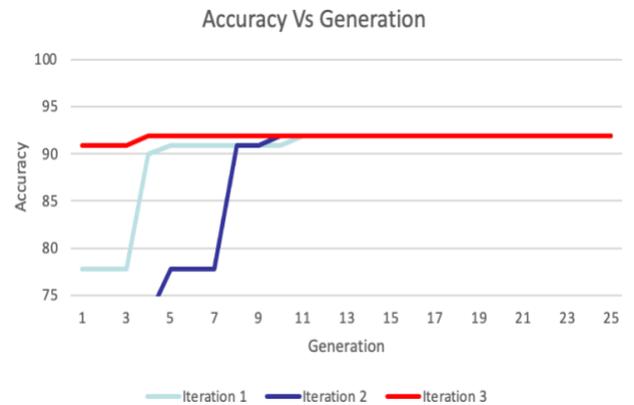

Fig. 9. Best value of fitness function across all the generation for three iterations

Table 2 shows the decoded chromosomes from the best candidate in each generation. In our experiments, AdamW and LAMB performed well, but Adafactor was never found in any of the elite candidate solutions. The highest accuracy with Adafactor was 63.7% for a learning rate=1e-03, epsilon=1e-8, and epochs=25, along with linear warmup and cosine annealing. We observed that training with Adafactor was also most time consuming, with a 3-fold increase in time for Adafactor compared to AdamW. LAMB found near optimal solutions and its time of training was better than that of AdamW for higher epochs. For instance, 50 epochs in LAMB using comparable learning rates and epsilon with AdamW, the LAMB optimizer finished the training 17 minutes faster than AdamW. The best accuracy was maintained by the model with AdamW until epoch 10, at the expense of training time compared to the model using the LAMB optimizer. The optimized model with the learned parameters, i.e., weights and biases, is stored using the Python torch module. PyTorch saves the model using the Pickle module with a .pt file extension. Our optimized model is available on GitHub [39].

TABLE 2. DECODED CHROMOSOMES WITH HIGHEST FITNESS FUNCTIONS ACROSS GENERATIONS

| Generation | Iteration 1 | Iteration 2 | Iteration 3 |
|---|---|---|---|
| 1 | LAMB, 50,lr = 0.00001,eps = 1e-06 | LAMB,25, lr = 0.00001,eps = 1e-05 | AdamW,10, lr=2e-7, eps =1e-07 |
| 2 | LAMB,50, lr = 0.00001,eps = 1e-05 | LAMB ,25,lr = 0.00001,eps = 1e-05 | AdamW,10, lr=2e-7, eps =1e-07 |
| 3 | LAMB,50, lr = 0.00001,eps = 1e-05 | LAMB,25, lr = 0.00001,eps = 1e-05 | AdamW,10, lr=2e-7, eps =1e-07 |
| 4 | LAMB ,25, lr = 0.001,eps = 1e-06 | LAMB ,25,lr = 0.00001,eps = 1e-05 | AdamW,10, lr=2e-8, eps=1e-08 |
| 5 | AdamW,20, lr=2e-5, eps =1e-08 | LAMB,50, lr = 0.00001,eps = 1e-05 | AdamW,10, lr=2e-8, eps=1e-08 |
| 6 | AdamW,20 ,lr=2e-6, eps =1e-08 | LAMB,50, lr = 0.00001,eps = 1e-05 | AdamW,10, lr=2e-8, eps=1e-08 |
| 7 | AdamW,15, lr=2e-6, eps =1e-08 | LAMB ,50,lr = 0.00001,eps = 1e-05 | AdamW,10, lr=2e-8, eps=1e-08 |
| 8 | AdamW,15, lr=2e-7, eps =1e-08 | AdamW,15, lr=2e-7, eps =1e-08 | AdamW,10, lr=2e-8, eps=1e-08 |
| 9 | AdamW,15 ,lr=2e-7, eps =1e-08 | AdamW ,15,lr=2e-7, eps =1e-08 | AdamW ,10,lr=2e-8, eps=1e-08 |
| 10 | AdamW,15, lr=2e-7, eps =1e-08 | AdamW ,10,lr=2e-8, eps=1e-08 | AdamW,10, lr=2e-8, eps=1e-08 |
| 11 | AdamW,15, lr=2e-8, eps=1e-08 | AdamW,10, lr=2e-8, eps=1e-08 | AdamW,10, lr=2e-8, eps=1e-08 |
| 12 | AdamW,10, lr=2e-8, eps=1e-08 | AdamW,10 lr=2e-8, eps=1e-08 | AdamW,10, lr=2e-8, eps=1e-08 |
| 13 | AdamW,10, lr=2e-8, eps=1e-08 | AdamW,10, lr=2e-8, eps=1e-08 | AdamW,10, lr=2e-8, eps=1e-08 |
| 14 | AdamW,10, lr=2e-8, eps=1e-08 | AdamW,10, lr=2e-8, eps=1e-08 | AdamW ,10,lr=2e-8, eps=1e-08 |
| 15 | AdamW,10, lr=2e-8, eps=1e-08 | AdamW,10, lr=2e-8, eps=1e-08 | AdamW,10, lr=2e-8, eps=1e-08 |
| 16 | AdamW,10, lr=2e-8, eps=1e-08 | AdamW,10, lr=2e-8, eps=1e-08 | AdamW,10, lr=2e-8, eps=1e-08 |
| 17 | AdamW,10, lr=2e-8, eps=1e-08 | AdamW,10, lr=2e-8, eps=1e-08 | AdamW,10, lr=2e-8, eps=1e-08 |
| 18 | AdamW,10, lr=2e-8, eps=1e-08 | AdamW,10, lr=2e-8, eps=1e-08 | AdamW,10, lr=2e-8, eps=1e-08 |
| 19 | AdamW,10, lr=2e-8, eps=1e-08 | AdamW,10, lr=2e-8, eps=1e-08 | AdamW,10, lr=2e-8, eps=1e-08 |
| 20 | AdamW,10, lr=2e-8, eps=1e-08 | AdamW,10, lr=2e-8, eps=1e-08 | AdamW,10, lr=2e-8, eps=1e-08 |
| 21 | AdamW,10, lr=2e-8, eps=1e-08 | AdamW,10, lr=2e-8, eps=1e-08 | AdamW ,10,lr=2e-8, eps=1e-08 |
| 22 | AdamW,10, lr=2e-8, eps=1e-08 | AdamW,10, lr=2e-8, eps=1e-08 | AdamW,10, lr=2e-8, eps=1e-08 |
| 23 | AdamW,10, lr=2e-8, eps=1e-08 | AdamW,10, lr=2e-8, eps=1e-08 | AdamW,10, lr=2e-8, eps=1e-08 |
| 24 | AdamW,10, lr=2e-8, eps=1e-08 | AdamW,10, lr=2e-8, eps=1e-08 | AdamW ,10,lr=2e-8, eps=1e-08 |
| 25 | AdamW,10, lr=2e-8, eps=1e-08 | AdamW,10, lr=2e-8, eps=1e-08 | AdamW,10, lr=2e-8, eps=1e-08 |

## VIII. DISCUSSION AND LIMITATION

The context of SDoH text in the clinincal notes is limited to a few sentences. In these situations its important to perform an informed search for hyperparameters without any manual intervention beyond the initial setup. We compared the hyperparameters obtained as part of this research with 1) hyperparameters used in the Clinincal BioBERT paper [23] and 2) hyperparameters in Han's paper [25] on multilabel classification of SDoH based on BERT. To compare all three sets of parameters, we trained all three models on our dataset. For the BERT model from [25], we observed significant overfitting with our dataset, hence we adjusted the dropout rate from 0.5 specifed in [25] to 0.7 to obtain the results in Table 4. Table 3 shows the hyperparameters and Table 4 shows the performance metrics.

TABLE 3. HYPERPARAMETERS OF THREE BASE CASES

| Our results | [23] | [25] |
|---|---|---|
| AdamW, learning rate=2e-08,epochs=10, epsilon=1e-08, batch size=16 | Adam,learning rate= 5-05, epochs=2/3/4, Epsilon=1e-12, batch size =16/32 | Adam with momentum=0.9, learning rate=1e-04, epochs=10, batch size=32 |

TABLE 4. PERFORMANCE METRICS OF ALL THREE MODELS

|  | Our results | [23] | [25] |
|---|---|---|---|
| Accuracy | 0.91919 | 0.8 | 0.5454 |
| Micro F1 score | 0.91919 | 0.8 | 0.5454 |
| Recall score | 0.833333 | 0.666 | 0.8333 |
| Presicion score | 1.0 | 1.0 | 0.555 |

We observed that our model was initially overfitting for some hyperparameter combinations, presumably due to the limited size of the training set. To avoid this, we adjusted the dropout ratio in the output layer.

The main limitation in this study is that not all of the 72,668 rows returned by the regex actually expressed an SDOH sentiment about *the patient*, even though all of them appeared to match a SOHO concept. Once we discovered this issue, we performed a manual review of a random sample of approximately 1500 rows. Of those, only 1054 rows expressed a powerful SDOH sentiment about the patient. In order to create a balanced dataset, we added another 1130 rows from the lab domain, which were "glaringly" irrelevant to SDOH.

## IX. CONCLUSION AND FUTURE WORK

We performed genetic algorithm-based hyperparameter tuning of a Clinical BioBERT model trained on SDoH data. Our analysis suggests the best configuration for the specific problem uses an AdamW optimizer with learning rate=2e-8, number of epochs=10 and epsilon=1e-08. This achieved an accuracy of 91.91% and minimal cross entropy loss. We also compared the optimal hyperparameters obtained by our research with hyperparameters in [23] (original Clinical BioBERT paper) and hyperparameters from [25] on SDoH text classification. We conclude that the hyperparameters obtained by informed search using the genetic algorithm outperformed the other models trained on the same dataset. The optimal hyperparameters presented in this paper for Clinical BioBERT should be tested with other clinical datasets, to determine if a similar accuracy improvement can be achieved for text classification.

The scope of improvement for this work is to use established techniques for ontology-based automatic annotation and human expert validation. In future work we will develop an annotated SDoH dataset from MIMIC III not limited to the social work category and determine whether the hyperparameters obtained in this paper can be used to improve accuracy with a larger dataset. Alternatively, a new set of hyperparameters will be found using the same genetic algorithm.